\documentclass[letterpaper, 10 pt, conference]{ieeeconf}
\IEEEoverridecommandlockouts
\usepackage{times}
\usepackage{soul}
\usepackage{url}
\usepackage[hidelinks]{hyperref}
\usepackage{graphicx}
\usepackage{amsmath}
\usepackage{amssymb}
\usepackage{booktabs}
\usepackage{algorithm}
\usepackage{algorithmic}
\usepackage{multirow}
\usepackage{array}
\usepackage{xcolor}
\usepackage{cite}
\usepackage{float}

\definecolor{purple}{RGB}{153, 0, 255}

\DeclareMathOperator*{\argmax}{arg\,max}

\usepackage{hhline}
\usepackage[caption=false]{subfig}

\begin{document}
\title{\LARGE \bf
Visually Grounded Task and Motion Planning for Mobile Manipulation
% Learning to Ground Spatial Relationships for \\Robot Task and Motion Planning
}
\author{Xiaohan Zhang$^1$, Yifeng Zhu$^2$, Yan Ding$^1$, Yuke Zhu$^2$, Peter Stone$^{2,3}$, Shiqi Zhang$^1$% <-this % stops a space
%\thanks{*This work was not supported by any organization}% <-this % stops a space
\thanks{$^1$~Department of Computer Science, The State University of New York at Binghamton \texttt{\{xzhan244; yding25; zhangs\}@binghamton.edu}}
\thanks{$^2$~Department of Computer Science, The University of Texas at Austin \texttt{\{yifeng.zhu@; yukez@cs.; pstone@cs.\}utexas.edu}}
\thanks{$^3$~Sony AI}
%}
}
%\thanks{$^1$Department of Computer Science, The State University of New York at Binghamton, NY 13902 \texttt{\{xzhan244; yding25; zhangs\}@binghamton.edu}, $^2$Department of Computer Science, The University of Texas at Austin, TX 78712 \texttt{\{yifeng.zhu@; yukez@cs.; pstone@cs.\}utexas.edu}, $^3$Sony AI. This work has taken place at the Autonomous Intelligent Robotics (AIR) Group, SUNY Binghamton. AIR research is supported in part by grants from the NSF (NRI-1925044), Ford, OPPO, and SUNY RF. Peter Stone serves as the Executive Director of Sony AI America.}}

\maketitle
\thispagestyle{empty}
\pagestyle{empty}

\begin{abstract}
Task and motion planning (TAMP) algorithms aim to help robots achieve task-level goals, while maintaining motion-level feasibility. 
This paper focuses on TAMP domains that involve robot behaviors that take extended periods of time (e.g., long-distance navigation). 
In this paper, we develop a visual grounding approach to help robots probabilistically evaluate action feasibility, and introduce a TAMP algorithm, called GROP, that optimizes both feasibility and efficiency. 
We have collected a dataset that includes $96,000$ simulated trials of a robot conducting mobile manipulation tasks, and then used the dataset to learn to ground symbolic spatial relationships for action feasibility evaluation. 
Compared with competitive TAMP baselines, GROP exhibited a higher task-completion rate while maintaining lower or comparable action costs. 
In addition to these extensive experiments in simulation, GROP is fully implemented and tested on a real robot system. 
\end{abstract}

% Robots frequently use task and motion planning (TAMP) algorithms to plan for long-horizon tasks while computing motion trajectories. 
% However, feasibility estimation at planning time is difficult due to robot imperfect perception capabilities at execution time.
% As a result, grounded TAMP algorithms aim to generate a feasible and efficient plan towards maximizing long-term utilities under real-world uncertainty.
% In this paper, we first define the problem \textbf{PMM} of perception-based TAMP for mobile manipulators.
% Then we propose an FCN-based method called GROP that for the first time optimizes task-motion plans by grounding spatial relationships.
% At the training phase, we collected a robot perception dataset including 92,000 data instances.
% At the evaluation phase, GROP has been quantitatively compared with competitive TAMP baselines in simulation and demonstrated on a real robot. 
% Experimental results show that our method can compute plans that are of high task-completion rates while maintaining relatively low action costs.
% \end{abstract}

\section{Introduction}
Task and motion planning (TAMP) algorithms and systems have been used for robot planning at both discrete and continuous levels~\cite{lagriffoul2018platform,garrett2021integrated}.
Task planners sequence symbolic actions for guiding the robot's high-level behaviors~\cite{ghallab2016automated}, and motion planners calculate low-level motion trajectories in continuous spaces~\cite{choset2005principles}. 
TAMP algorithms aim to bridge the gap between task planning and motion planning towards enabling robots to fulfill task-level goals and maintain motion-level feasibility at the same time~\cite{toussaint2015logic, zhu2020hierarchical, garrett2018ffrob, jiang2019task, thomas2021mptp, wells2019learning, migimatsu2020object, mcmahon2017robot, zhao2018reactive}.

One way to categorize TAMP domains is based on if a problem domain requires robot actions that take relatively short time (e.g., seconds, such as picking up and putting down objects) or relatively long time (e.g., minutes or even hours as navigating from one location to another)~\cite{lo2020petlon}. 
In the former type of domains, action feasibility is much more important to consider than plan efficiency, since extra plan steps do not add much time to the expected execution time.
On the other hand, this paper is motivated by the latter type of TAMP domains, wherein it is advantageous to incorporate both \textit{efficiency} and \textit{feasibility} into the evaluation of plan qualities.
Some existing TAMP research incorporates both efficiency and feasibility into task-motion planning~\cite{lo2020petlon,thomas2021mptp}. 
% However, those methods that consider both efficiency and feasibility require predefined ``state mapping functions''\cite{lo2020petlon} for mapping each symbolic state into one feasible pose in continuous space. 
However, those methods evaluate feasibility in a deterministic way, and rely on predefined ``state mapping functions'' for mapping each symbolic state into feasible poses in continuous space. 
For instance, to unload an object to a table, a robot needs to move to the table first, i.e., \texttt{beside(table)=true}, where previous TAMP research that we are aware of relies on predefined feasible poses that are spatially close to the table to evaluate the truthfulness of the ``beside table'' statement.

Such predefined state mapping functions that assume deterministic action feasibility have at least two deficiencies.
%There are at least the following two issues from the state mapping functions that are predefined and assume deterministic feasibility evaluation. 
First, a predefined state mapping function is not robust to dynamic obstacles (e.g., people seated around the table). 
Second, not all ``feasible'' behaviors are equally preferred, e.g., standing far and stretching out to place an object may be less preferred than standing close to do so.
Those observations motivate this work that learns to evaluate action feasibility for robot task-motion planning.

\begin{figure}
\begin{center}
    \vspace{.7em}
    \includegraphics[width=0.4\textwidth]{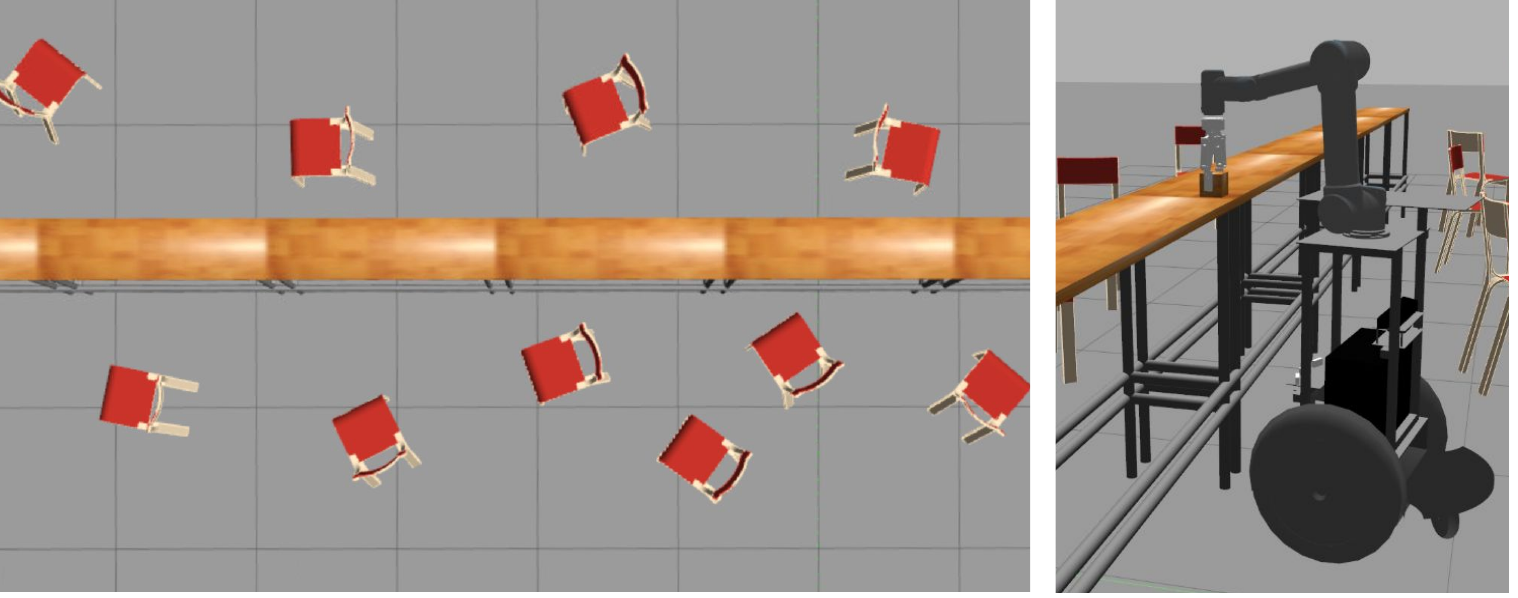}
    \vspace{-.5em}
    \caption{Our mobile manipulation domain that includes a long banquet table surrounded by chairs. 
    Given a target location (on the table) to place an object, the robot needs to navigate to a location from which it can successfully perform the manipulation action, ideally as quickly as possible (thus preferring the near side of the table when feasible).}
    % \caption{Everyday objects used in this research~\cite{thomason2016learning}, and the robot arm used for performing exploratory actions (grasping an object). }
    \vspace{-2em}
    \label{fig:serving}
\end{center}
\end{figure}

%\xiaohan{Do we still want to mention feasibility evaluation as one of our contribution?}
The main contribution of this work is a visually grounded TAMP algorithm, called \textbf{G}rounded \textbf{RO}bot Task and Motion \textbf{P}lanning (GROP), that probabilistically evaluates action feasibility, and incorporates both feasibility and efficiency towards maximizing long-term utility. 
Inspired by the concept of ``symbol grounding''~\cite{harnad1990symbol}, we use ``visual grounding'' to refer to methods that use computer vision techniques to help an agent interpret abstract symbol tokens and connect them to the real world. 
%\yifeng{from visual observations? I think it is a little bit ambiguous to understand what does it mean by "interpreting symbols" here.}
% The motion-level feasibility values are incorporated into task planning, enabling our robot to dynamically decide when and how to grasp objects (e.g., from which side of a table). 

\begin{figure*}
%https://docs.google.com/drawings/d/1pjLwngNUAq8LNlY1_8Yae1kUuBnjzRO_4LFAL8Cd8vQ/edit
\begin{center}
    %\vspace{.5em}
    % \includegraphics[width = 17.5em,height=9em]{Figures/objects.png}
    \includegraphics[width=0.95\textwidth]{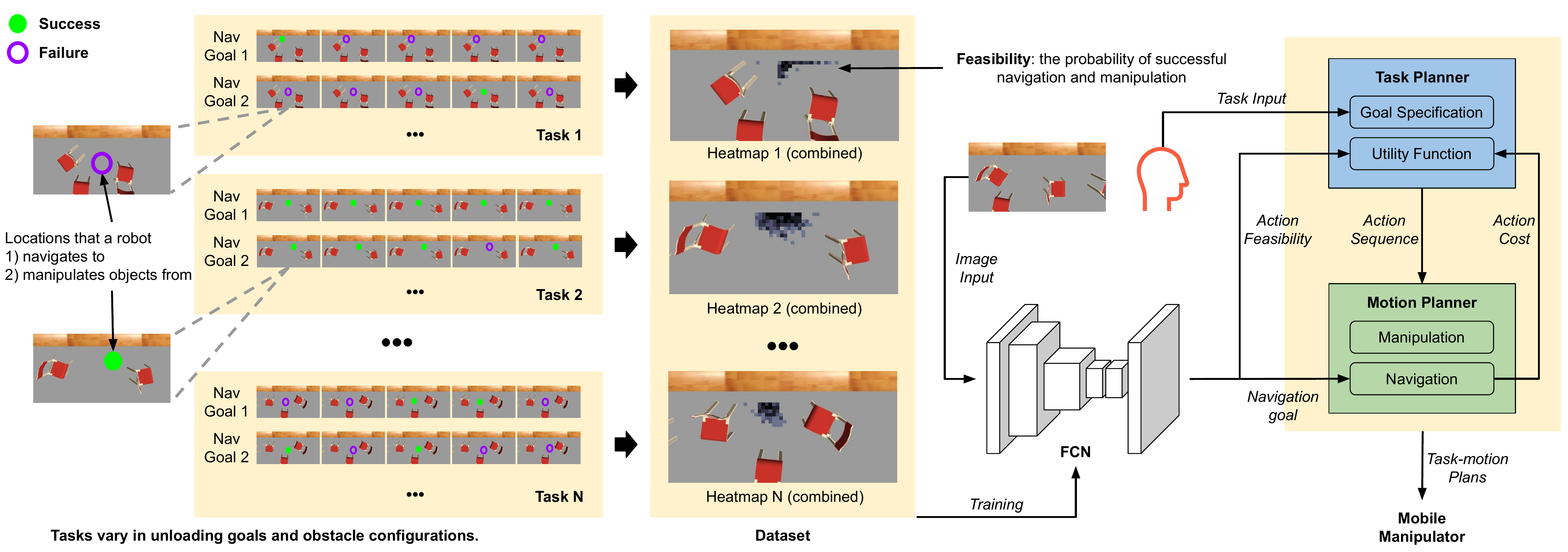}
    \vspace{-.5em}
    \caption{An overview of this work, including an FCN-based feasibility evaluation approach, and GROP, our grounded TAMP algorithm.
    A \emph{task} corresponds to one ``unloading goal'' on the table, as well as a configuration of obstacles (chairs in our case). 
    Given a task, every pixel is considered a navigation goal -- the robot attempts to navigate there, and unload an object from there. 
    This navigation-manipulation process is referred to as a \emph{trial}. 
    The robot performs multiple trials for each navigation goal, which yields a \emph{feasibility} value for that particular location. 
    The feasibility values together form one \emph{heatmap} for each task. 
    In our \emph{dataset}, each instance is a top-down view image, whose label is the corresponding heatmap. 
    The ``Dataset'' box shows a few ``combined heatmaps'' where heatmaps are overlaid onto the corresponding images. 
    Training with the dataset generates an FCN that is used for two purposes: 1)~evaluating the feasibility of task-level actions, and 2) selecting motion-level navigation goals. 
    Finally, GROP incorporates both efficiency (measured by action costs) and feasibility to compute task-motion plans for a mobile manipulator. 
    } 
    \vspace{-2em}
    \label{fig:overview}
\end{center}
\end{figure*}

We have applied GROP to a domain of a mobile manipulator setting ``dinner tables,'' as illustrated in Fig.~\ref{fig:serving}. 
The robot needs to decide how to approach a table at the task level (e.g., from which side of the table), compute 2D navigation goals (connecting task and motion levels), and plan motion trajectories for navigation and manipulation behaviors. 
We have collected a dataset that includes $96,000$ instances of a robot conducting mobile manipulation tasks where in each instance, a robot unloads an object with dynamic obstacles surrounding a table.
%\yifeng{Maybe directly saying "dinner tables".}
An instance is labeled ``successful'' if the robot is able to compute and execute a task-motion plan that includes both navigation and manipulation actions. 
We use fully convolutional networks (FCNs)~\cite{long2015fully} to learn to visually ground spatial relationships and evaluate action feasibility. 
GROP is summarized in Fig.~\ref{fig:overview}. 

Compared with baselines from the literature~\cite{lo2020petlon,driess2020deeph}, GROP performed better in success rate while maintaining lower (or comparable) cumulative action costs.
Finally, we demonstrate GROP with real-world robot hardware.

% In current TAMP systems, to plan navigation and manipulation behaviors, task planners use rules that given as prior knowledge to the agent. 
% For instance, an example rule can state ``\textit{A robot is able to unload an object to a table if the robot is \textbf{beside} the table}''.
% %However, evaluating the value of the ``beside'' predicate is difficult due to factors such as nearby obstacles and table shapes. 
% In this case, the spatial relationship such as ``beside'' predicate is very important for evaluating if the unloading behavior will be successful.
% However, the value of the ``beside'' predicate is difficult to estimate because the robot cannot accurately perceive nearby obstacles and the previous navigation action causes small-scale uncertain outcomes. 
% %What makes it more challenging is the agents' imperfect perception capabilities (figure~\ref{fig:perception}) and uncertain action outcomes. 
% As a result, we propose a method called GROP to visually ground spatial relationships towards efficient and feasible TAMP under real-world uncertainty. 
% We first collected a pixel-wise robot perception dataset in simulation containing 100 scenes (Section~\cite{}). 
% Then we trained a robot perception module using fully convolutional network (FCN) to generate unloading heatmaps for guiding TAMP (Section~\cite{}). 
% We have extensively evaluated GROP in simulation and results show that our method outperforms several competitive TAMP baselines. 

%Our contributed algorithm

\section{Related Work}
TAMP methods aim to compute plans that fulfill task-level goals while maintaining motion-level feasibility, as reviewed in recent articles~\cite{lagriffoul2018platform,garrett2021integrated}. 
Several TAMP algorithms have been introduced in recent years (e.g.,~\cite{gravot2005asymov,plaku2007discrete,erdem2011combining,srivastava2014combined,lagriffoul2014efficiently, chitnis2016guided,garrett2018ffrob,wang2018active,kim2019learning,chitnis2019learning,ding2022learning,zhu2020hierarchical,kim2020learning, dantam2018incremental}.
Within the TAMP context, we distinguish a few subareas of TAMP that are closest to this research on learning to visually ground symbolic spatial relationships towards planning efficient and feasible task-motion behaviors under uncertainty. 

\subsection{TAMP for Efficient and Feasible Behaviors}
When high-level actions only take a few seconds, TAMP algorithms can focus mostly on action feasibility constraints without fully optimizing high-level plan efficiency.
%\yifeng{executing actions take ...} 
However, when there are actions that take significant time to execute (e.g., long-distance navigation), task-completion efficiency cannot be overlooked.
Some recent methods have considered efficiency in different aspects of TAMP, such as planning task-level optimal behaviors in navigation domains~\cite{lo2020petlon}, integrating reinforcement learning with symbolic planning in dynamic environments~\cite{jiang2019task}, computing safe and efficient plans for urban driving~\cite{ding2020task}, and optimizing robot navigation actions under the uncertainty from motion and sensing~\cite{thomas2021mptp}.
In contrast to those methods that do not have a perception component, GROP visually grounds symbols (about spatial relationships) to probabilistically evaluate action feasibility for task-motion planning.

\subsection{TAMP under Uncertainty}
While most TAMP methods assume a fully observable and deterministic world~\cite{garrett2021integrated}, some have been developed to account for the uncertainty from perception and action outcomes~\cite{kaelbling2013integrated,hadfield2015modular,phiquepal2019combined,garrett2020online, nouman2021hybrid, akbari2020contingent}. 
For instance, the work of Kaelbling and Lozano-P{\'e}rez extended the ``hierarchical planning in the now'' approach to address both current-state uncertainty and future-state uncertainty~\cite{kaelbling2013integrated}. 
Going beyond those methods that aim to maintain plan feasibility to complete tasks under high-level uncertainty, we consider uncertainty in the robot motion and also incorporate task-completion efficiency into the optimization of robot behaviors. 
As a result, our GROP algorithm is particularly suitable for TAMP domains that require robot operations over extended periods of time, such as long-distance navigation. 
% (mostly from navigation tasks) that differs in the degree to which it is important to optimize the task plan.

%\shiqi{It's very hard to justify that we ``more focused'' on something. We shall find a feature of this work that is absolutely new. If none of those methods considered efficiency, then we shall just say it that way. }
%\xiaohan{This work~\cite{thomas2021mptp} also consider navigation costs under uncertainty. But they have predefined state-mapping function. Should we discuss this work in details?}

\subsection{TAMP with Visual Perception}
%\shiqi{We should first say TAMP agent, by default, does not learn. }
%Recently, diverse learning methods have been used to speed-up searching process in current TAMP systems~\cite{zhu2020hierarchical,kim2020learning,wang2018active,chitnis2016guided,kim2019learning,chitnis2019learning}.
%Our work focuses on using deep neural networks to learn motion-level feasibility.
Recently developed methods have shown that visual information can be used to help robots predict plan feasibility, including task-level feasibility~\cite{driess2020deepr,zhu2020hierarchical}, and motion-level feasibility~\cite{driess2020deeph,wells2019learning}. 
Those methods were developed to maximize task completion rate in manipulation domains, and actions that take relatively long time (such as long-distance navigation) were not included in their evaluations. 
Focusing on robots that operate over extended periods of time, GROP (ours) incorporates efficiency into plan optimization. 
For instance, when highly feasible plans have very high costs, GROP supports the flexibility of executing slightly less feasible plans with much lower costs. 
%We attribute this desirable trade-off between feasibility and efficiency to GROP's capability of probabilistically evaluating plan feasibility, which is not supported by the above-mentioned methods.
GROP achieves this desirable trade-off between feasibility and efficiency by probabilistically evaluating plan feasibility, which is not supported by the above-mentioned methods. 
%However, they all assumed that feasibility is a deterministic value, which means that there is no gap between planning time and execution time.
%\shiqi{I don't understand how this statement connects to learning. }
%To our knowledge, GROP is the first that learns evaluated feasibility under real-world uncertainty.

\section{Problem Statement}
\label{sec:problem_statement}

We consider a mobile manipulation domain that includes $N$ objects $\textit{Obj}$. 
There are obstacles (tables and chairs in our case) that prevent the robot from navigating to some positions in the domain. 
Location $l$ is a symbolic concept that corresponds to a set of obstacle-free 2D poses ($X$), where each pose ($x\in X$) specifies a 2D position and an orientation. 
The robot needs to move each object $o \in \textit{Obj}$ from its initial location to a goal position. 

\vspace{.5em}
\noindent
\textbf{Actions: }The robot is equipped with skills of performing a set of symbolic (task-level) actions denoted as $A: A^n \cup A^m$, where $A^n$ and $A^m$ are \emph{navigation} actions and \emph{manipulation} actions respectively.
A navigation action $a^n_{l, l'} \in A^n$ is specified by its initial and goal locations, $l, l'\in L$, where $L$ includes a set of symbolic locations. 
A manipulation action, $a^m_{o, l} \in A^m$, is specified by an object to be manipulated, $o \in \textit{Obj}$, and a symbolic location, $l \in L$, to which the robot navigates and performs the manipulation action. 
We consider two types of manipulation actions of loading and unloading, represented by $a^{m+}$ and $a^{m-}$ respectively. 
%Actions generally have preconditions and effects.
%For instance, the action \texttt{load($o_1$)} (where \texttt{object($o_1$)}, \texttt{location($l_1$), \texttt{atLocation($o_1$,$l_1$)}}) requires preconditions such as \texttt{at(rob,$l_1$)}, and results in effects such as \texttt{inHand($o_1$)}.
Actions are defined via preconditions and effects. 
For instance, the action \texttt{load($o_1$)} has preconditions of \texttt{at(robot,$l_1$)} and \texttt{at($o_1$,$l_1$)}, meaning that to load the object $o_1$, the object must be co-located with the robot at the location $l_1$. 
The effects of \texttt{load($o_1$)} include $o_1$ being moved into the robot’s hand, i.e., \texttt{inhand($o_1$)}. 

\vspace{.5em}
\noindent
\textbf{Perception: }
The robot visually perceives the environment through top-down views over the areas where manipulation and navigation actions are performed. 
We use $\textit{IM}$ to represent a 2D image that captures the current obstacle configuration, as shown in the ``Image Input'' of Fig.~\ref{fig:overview} (bottom right). 
To facilitate robot learning, we provide a dataset (as illustrated in the ``Dataset'' box of Fig.~\ref{fig:overview}). 
Each instance includes a top-down view image, and a target object with a predefined position, while each label is in the form of a heatmap. 
Each pixel of a heatmap is associated with a 2D position, and has a ``feasibility'' value that represents the success rate of the robot navigating to the 2D position, and manipulating the target object from there. 

A map is generated in a pre-processing step, and provided to the robot as prior information for navigation purposes using rangefinder sensors. 
%Different $M$ may lead to different planning results both at task and motion levels

\vspace{.5em}
\noindent
\textbf{Uncertainty: }
The outcome of performing navigation action $a^n_{l,l'}$ to goal pose $x$ is deterministic at the task level, but is non-deterministic at the motion level. 
In other words, the robot will end up in position $x'$, which is not necessarily the same as $x$. 
This setting captures the fact that a mobile robot never achieves its exact 2D navigation goal (due to its imperfect localization and actuation capabilities), though successfully navigating to an area ($l$) is generally possible. 

We focus on the interdependency between navigation and manipulation actions. 
For instance, the execution-time uncertainty from navigation actions results in different standing positions of the robot, which makes the outcomes of manipulation actions non-deterministic. 
%\yifeng{I think if we can explain why uncertainty in execution time leads to different standing positions, that would be better (I assume this means execution time -> imperfect trajectory following to navigation planner?)}
This challenge generally exists in mobile manipulators. 
We assume no noise in the execution of manipulation actions (loading and unloading) to objects within a reachable area. 

\vspace{.5em}
\noindent
\textbf{Format of Solution: }
% Within a task-motion planning context, each object's task-level goal location $l$ is associated with a motion-level position $x$, and $x\in f(l)$. 
A solution is in the form of a task-motion plan 
$
    p=\langle p^t, p^m \rangle, 
$
where task plan $p^t$ is of the form $\langle a_0^n, a_0^m, a_1^n, a_1^m, ...\rangle$, indicating that navigation and manipulation actions are interleaved. 
% , where $a_k^n$ is a symbolic navigation action (of \texttt{goto}) and $a_k^m$ is a symbolic manipulation action (of \texttt{load} and \texttt{unload}).
Motion plan $p^m$ is of the form $\langle \xi_0^n,\xi_0^m, \xi_1^n,\xi_1^m, ... \rangle$, and $\xi_i^n$ (or $\xi_i^m$) is a trajectory in continuous space for implementing symbolic action $a_i^n$ (or $a_i^m$).
The quality of task-motion plan $p$ is evaluated using a utility function $\mathcal{U}(p)$, which considers both feasibility and efficiency of plan $p$:
\begin{align}
   \mathcal{U}(p) = \mathcal{R} \cdot \mathcal{F}(p) - \mathcal{C}(p),
\label{eqn:utility}
\end{align}
where $\mathcal{F}(p) \in [0,1]$ is the plan feasibility (i.e., the probability that $p$ can be successfully executed), $\mathcal{C}(p)$ is the overall plan cost of executing $p$, and $\mathcal{R}\!\!\!\rightarrow \!\!\!\mathbb{R}$ is a success bonus reflecting the reward from a successful execution. 
An optimal algorithm reports a task-motion plan of the highest utility: 
$$
    p^* = \argmax_{p} ~\mathcal{U}(p) 
$$

Next, we present an algorithm that computes such task-motion plans through visually grounding spatial relationships while considering both efficiency and feasibility.

%A robot is given a task where we denote the set of goal specifications as $G$.
%For example, to unload two objects $obj_1$ and $obj_2$ to two different locations $loc_1$ and $loc_2$, $G$ includes \texttt{unloadedto($obj_1$, $loc_1$)} and \texttt{unloadedto($obj_2$, $loc_2$)}.
%Let $I$ be the set of initial configurations including all domain variables (e.g., robot and environments).. 

\section{The GROP Algorithm}
\label{sec:alg}

In this section, we introduce the paper's main contribution, an algorithm called \textbf{G}rounded \textbf{RO}bot Task and Motion \textbf{P}lanning, or GROP for short. 

\subsection{Algorithm Description}

%Aiming at maximizing the total utility, we introduce a new term called \textbf{spatial feasibility}.
%The requirements of GROP include a task planner $P^t$, a motion planner $P^m$, reward $\mathcal{R}$, robot model $M$, function $h$ (Equation~\ref{eqn:task-mapping}), and a trained feasibility evaluator $\Psi$.
Algorithm~\ref{alg:GROP} presents the GROP algorithm. 
Implementing GROP requires a task planner $\textit{Plnr}^t$, a motion planner $\textit{Plnr}^m$, a success bonus $\mathcal{R}\!\!\rightarrow \!\!\mathbb{R}$, and a cost function $\textit{Cst}$ that evaluates the cost of any motion trajectory generated by $\textit{Plnr}^m$. 
Inputs of GROP include a rule-based task description $T$, a robot initial 2D position $x^{\textit{init}}$, and a provided dataset $D$.
GROP outputs a task-motion plan $p$ in the form of $\langle p^t, p^m \rangle$.

%GROP starts with generating task-level areas $G$ using $T$ and function $h$ (Line 1).
%Function $f$ (Equation~\ref{eqn:motion-mapping}) is initialized to include all the feasible and infeasible poses in the navigation domain (Line 2).
GROP starts with training an FCN-based feasibility evaluator $\Psi$ using provided dataset $D$ in Line~\ref{l:train}. 
Then it initializes an empty set of task-motion plans $\textbf{P}$ in Line~\ref{l:initP}.
$\textit{Plnr}^t$ takes $T$ as input and outputs a set of task-level satisficing plans, denoted as $\textbf{P}^t$ in Line~\ref{l:task}.
The outer for-loop (Lines~\ref{l:oloops}-\ref{l:oloope}) iterates over each task-level satisficing plan.
In each iteration, GROP evaluates the utility value of one task plan $\mathcal{U}(p)$, which incorporates both plan feasibility $\mathcal{F}(p)$ and plan efficiency $\mathcal{C}(p)$.
% The outer loop iterates over each task plan to evaluate its utility value $\mathcal{U}(p)$. 
%$\mathcal{U}(p)$ takes both plan feasibility $\mathcal{F}(p)$ and plan efficiency $\mathcal{C}(p)$ into account.
Aiming to evaluate $\mathcal{F}(p)$ and $\mathcal{C}(p)$, each iteration in the first inner for-loop (Lines~\ref{l:iloops}-\ref{l:iloope}) considers a pair of navigation and manipulation actions in the task plan, and evaluates its feasibility and cost.
In the second inner for-loop of Lines~\ref{l:motions}-\ref{l:motione}, GROP calls $\textit{Plnr}^m$ to compute one motion plan for each task-level action.
% The computed task plan $p^t$ and motion plan $p^m$ together form a task-motion plan $p$, which is then added to set $\textbf{P}$ (Line~\ref{l:tmp}).
Line~\ref{l:tmp} puts together task plan $p^t$ and motion plan $p^m$ to form a task-motion plan $p$.
In the same line, $p$ is added to task-motion plan set $\textbf{P}$.
Lines~\ref{l:select}-\ref{l:return} are the final steps to select and return the optimal task-motion plan from $\textbf{P}$ given utility function $\mathcal{U}(p)$.

%GROP computes task-motion plans by incorporating both feasibility and efficiency into the optimization process. 
%One important step of GROP is to evaluate task-level action feasibility (Line~\ref{l:feas}), which is discussed next. 

\begin{algorithm}[t]
\caption{GROP}\label{alg:GROP}  \small
\begin{algorithmic}[1]
\REQUIRE Task planner $\textit{Plnr}^t$, motion planner $\textit{Plnr}^m$, success bonus $\mathcal{R}$, and cost function $\textit{Cst}$\\
\hspace{-1.7em}\textbf{Input:} Task description $T$, robot initial position $x^{\textit{init}}$, dataset $D$
\STATE {Train a motion-level feasibility evaluator $\Psi$ using dataset $D$ (detailed in Section~\ref{sec:feasibility})} \label{l:train}
\STATE {Initialize a set of task-motion plans $\textbf{P} \leftarrow \emptyset$} \label{l:initP}
%\STATE {Initialize $L$ given task description $T$}
%\STATE {Initialize $f(l)$ with all obstacle-free poses in the domain for each $l \in L$}
\STATE {Compute a set of task-level satisficing plans: $\textbf{P}^t\leftarrow \textit{Plnr}^t(T)$} \label{l:task}
%\STATE {Initialize $\mathcal{C}(p^t) \leftarrow 0$ and $\mathcal{F}(p^t) \leftarrow 0$ for each $p^t \in \textbf{P}^t$}
\FOR {each plan $p^t \in \textbf{P}^t$} \label{l:oloops}
    \STATE {Initialize a motion-level position sequence: $X^{\textit{seq}} \leftarrow [x^{\textit{init}}]$}
    \STATE {Initialize $\textit{tmp}^f \leftarrow 0$ and $\textit{tmp}^c \leftarrow 0$}
    \FOR {each action pair $\langle a^n_{l, l'}, a^m_{o, l'} \rangle$ in $p^t$} \label{l:iloops}
%\FOR {each action pair $\langle a^n, a^m \rangle$ in $p$}
        \STATE {Capture $\textit{IM}$ of location $l'$}
        \STATE {Predict heatmap $h = \Psi(\textit{IM})$, using Eqn.~\ref{eqn:heatmap}}
 %       \STATE {Evaluate task-level feasibility $Fea^t(a^n_{l, l'}, a^m_{o, l'} )$, using Eqn.~\ref{eqn:task_feasibility}}
        \STATE {$\textit{tmp}^f \leftarrow \textit{tmp}^f + \textit{Fea}^t(a^n_{l, l'}, a^m_{o, l'} )$, using Eqn.~\ref{eqn:task_feasibility}} \label{l:feas}
        \STATE {$x' \leftarrow \textit{Smp}( l',h )$, and append $x'$ to $X^{\textit{seq}}$} \label{l:smp}
        \STATE {$\textit{tmp}^c \!\! \leftarrow \!\! \textit{tmp}^c + \textit{Cst}\big(\textit{Plnr}^m(a^n_{l, l'})\big) + \textit{Cst}\big(\textit{Plnr}^m(a^m_{o, l'})\big)$ }
    \ENDFOR \label{l:iloope}
%\STATE {$\mathcal{F}(p) \leftarrow \sum_{\langle a^n_{l, l'}, a^m_{o, l'} \rangle \in p}Feasibility(a^n_{l, l'}, a^m_{o, l'})$}
%\STATE {Evaluate plan feasibility $\mathcal{F}_p$ using $IM$}
%\STATE {$\mathcal{C}(p) \leftarrow \sum_{\langle a^n_{l, l'}, a^m_{o, l'} \rangle \in p}Cost(a^n_{l, l'}, a^m_{o, l'})$}
%\STATE {Evaluate total action costs $\mathcal{C}_p$}
    \FOR{each $(x_i, x_{i+1}) \in X^{\textit{seq}}$} \label{l:motions}
        \STATE {Compute motion-level trajectory $\xi \leftarrow P^m(x_i, x_{i+1})$}
        \STATE {Append $\xi$ to motion plan $p^m$}
    \ENDFOR \label{l:motione}
    \STATE {Generate task-motion plan $p \leftarrow \langle p^t, p^m \rangle$, and append $p$ to the task-motion plan set $\textbf{P}$} \label{l:tmp}
    %\STATE {Append $p$ to $\textbf{P}$}
    \STATE {Update $\mathcal{F}(p) \leftarrow \textit{tmp}^f$ and $\mathcal{C}(p) \leftarrow \textit{tmp}^c$}
    \STATE {$\mathcal{U}(p) \leftarrow \mathcal{R} \cdot \mathcal{F}(p) - \mathcal{C}(p)$ (Eqn.~\ref{eqn:utility})}
\ENDFOR \label{l:oloope}

\STATE {Compute optimal task-motion plan: $p^* = \argmax_{p \in \textbf{P}} \mathcal{U}(p)$} \label{l:select}

% \STATE {Compute trajectories for $X^{seq}$ using $P^m$ and update motion plan $p^m$}
\RETURN {$p^*$} \label{l:return}

\end{algorithmic}
\end{algorithm}

%\vspace{-3em}

%\subsection{Motion Planning}
\subsection{Feasibility Evaluation}
\label{sec:feasibility}

In this subsection, we discuss how to evaluate action feasibility at task and motion levels (Line~\ref{l:feas} in Algorithm~\ref{alg:GROP}), where the feasibility evaluation at the task level relies on the feasibility evaluation at the motion level. 

%\vspace{-10em}
\noindent
\textbf{Motion-Level Feasibility: }
In our mobile manipulation domain, motion-level feasibility $\textit{Fea}^m(x,y)$ is a function of 2D positions $x$ and $y$, and is the probability of a robot successfully navigating to $x$ and manipulating an object that is in position $y$. 
$\textit{Fea}^m(x, y)$ can be extracted from gray-scale heatmap image $h^y$ that is centered around $y$: 
\begin{align}
    \textit{Fea}^m(x,y) = h^y[x]
\end{align}

We use a FCN-based feasibility evaluator $\Psi$ to generate heatmap $h^y$, given a top-down view image $\textit{IM}^y$ captured right above unloading position $y$ (``Image Input'' in Fig.~\ref{fig:overview}): 
\begin{align}
    h^y = \Psi(\textit{IM}^y)
\label{eqn:heatmap}
\end{align}

% \vspace{.5em}
\noindent
\textbf{Data Collection and Learning $\Psi$ with FCN: }
Here we discuss how to learn $\Psi$ in Equation~\ref{eqn:heatmap}. 
% We first describe our data collection process, and then describe our neural network architecture for training. 
A \emph{task} specifies an obstacle configuration and a position $y$ that a robot wants to unload objects to. 
In each \emph{trial} of our data collection process, a robot attempts to navigate to position $x$, and then unload an object to position $y$. 
Such a trial produces a data point in the following format: 
$$
    (\textit{IM}^y, x):r
$$
where $\textit{IM}^y$ is a top-down view image captured right above $y$, and $r$ is either $\textit{true}$ or $\textit{false}$ depending on if the robot succeeds in both navigation and manipulation actions. 
%Due to the uncertainty of the unloading behavior in the real-world execution, unloading results $r$ may vary even for the same image and robot pose. 
The robot repeated the same process for $N$ times ($N=5$ in our case), and we used the results ($r_0, r_1, \cdots, r_{N-1}$) to compute a success rate for positions $x$ and $y$, which determines a gray-scale color for one pixel of a heatmap: $h[x]$. 
% Mathematically, $h[x]$ is computed as follows: 
% \begin{align}
%     h[x] = \frac{|\{ r \in R^{IM}_x, r = true \}|}{|R^{IM}_x|}
% \end{align}

Iterating over all possible positions of $x$ in an area of $\textit{Width} \times \textit{Height}$ ($24$ pixels by $8$ pixels in our case) in image $\textit{IM}$, we were able to generate one full heatmap $h$ for the current task. 
Here we assume this area is large enough to cover all positions, from which the robot can unload objects to $y$. 
% Let $R^{IM}_x$ be the set of unloading results from different trials.
% We convert $R^{IM}_x$ into a gray-scale color pixel (range in [0, 255]) and integrate many those pixels into a gray-scale color heatmap as a label of $IM$:
To diversify the instances, we randomly placed obstacles (chairs in our case) to generate ten different ``environments,'' and then randomly sampled unloading positions to generate a total of 100 tasks. 
As a result, our dataset contains 100 instances, each in the form of a top-down view image ($64\times 32$). 
Each instance has a label that is in the form of a heatmap. 
The size of our dataset is $96,000$, i.e., $100\! \times \! N \! \times \! \textit{Width}\!  \times \! \textit{Height}$. 

% There are 100 images (scaled into 64x32 images) being generated to train our FCN model.
% Along the dataset collection procedure, the robot was tasked to navigate to an unloading position corresponding to a pixel point, and perform an unloading behavior.
% Each pixel point was evaluated for 5 times
% We then labeled the data point (a partial view and an unloading position) by the unloading result (successful or not).
% Running a full simulation to collect the pixel-wise dataset is time-consuming, so we did two adjustments.
% One is teleporting the robot directly to the navigation goal where we use 2D Gaussian distribution to model the uncertainty of navigation action outcomes.
% The other one is only letting the robot explore an area near the table which contains 24x8 pixels.

\vspace{.5em}
\noindent
\textbf{Task-Level Feasibility: }
$\textit{Fea}^t(a^n_{l, l'}, a^m_{o, l'})$ evaluates the feasibility (in the form of a probability) of a robot successfully performing both task-level navigation action $a^n_{l, l'}$ and task-level manipulation action $a^m_{o, l'}$. 
\begin{align}
    \textit{Fea}^t(a^n_{l, l'}, a^m_{o, l'}) = \frac {\sum\limits_{i=0\cdots N-1} \!\!\! \textit{Fea}^m\big( \textit{Smp}_i(l',h), y \big) }{N}
\label{eqn:task_feasibility}
\end{align}
where function $\textit{Smp}_i(l', h)$ samples the $i$th 2D position from location $l'$. The positions are weighted by heatmap $h$ that is centered around object $o$. 
Intuitively, positions of higher motion-level feasibility are more likely to be sampled.

\section{Experiments}
We conducted extensive experiments in simulation, where a mobile manipulator performs navigation and manipulation actions to set ``dinner table.'' 
We also demonstrate GROP using a real robot system that includes a mobile platform and a robot arm. 
Our main hypothesis is that GROP outperforms existing TAMP algorithms in task completion rate without introducing additional action costs. 
%\shiqi{Literally, it does not make sense to compare a hypothesis to baselines. This sentence should be rephrased. }

\begin{figure*}[t]
    \centering
    %  \hfill
    %\subfloat[Target positions]{\includegraphics[width=.24\linewidth]{figures/fig_original.pdf}}\qquad \hspace{-1.7em} 
    \vspace{-2em}
    \subfloat[Easy]{\includegraphics[width=.32\linewidth]{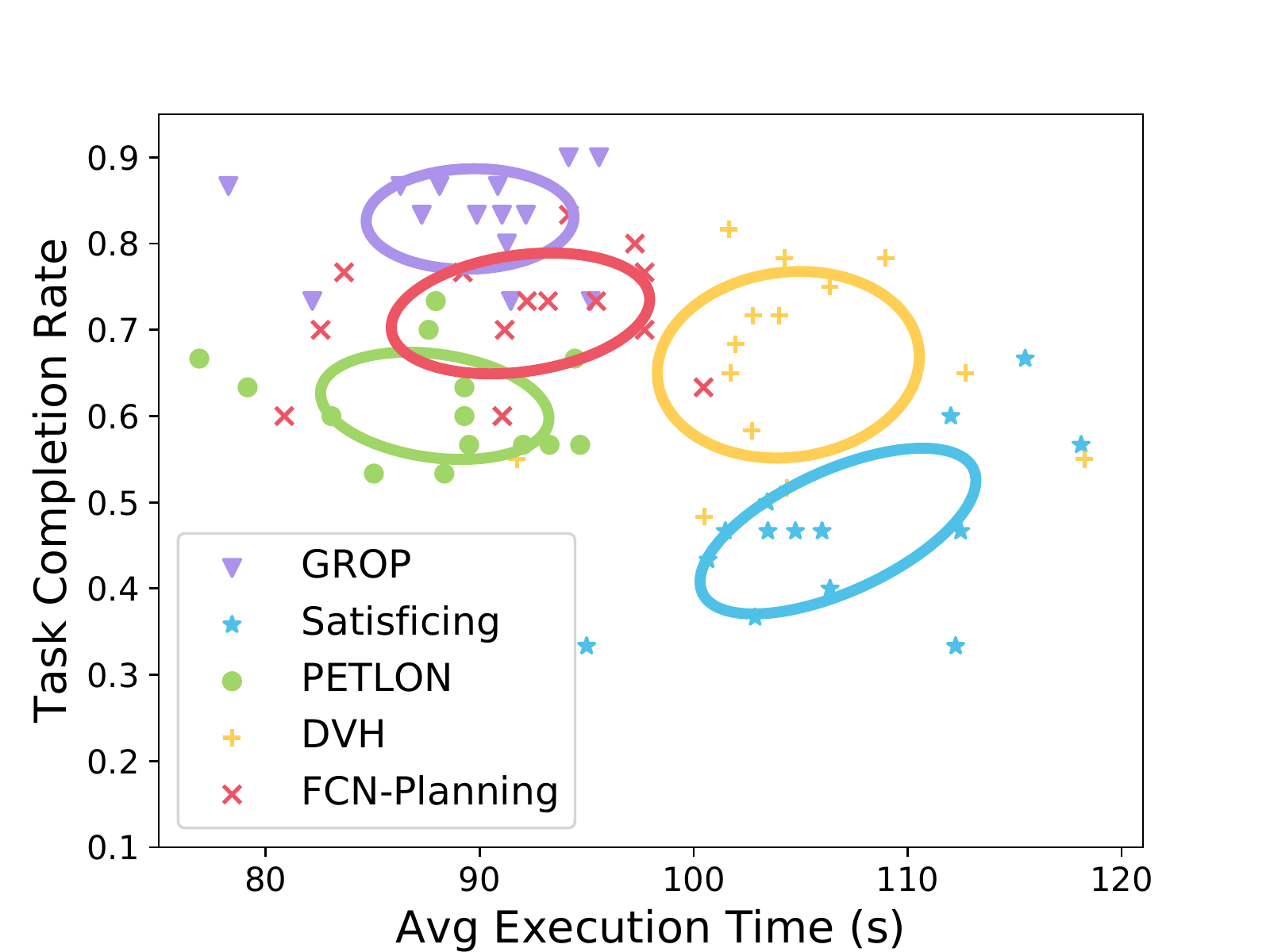}}\qquad \hspace{-2em}
    \subfloat[Normal]{\includegraphics[width=.32\linewidth]{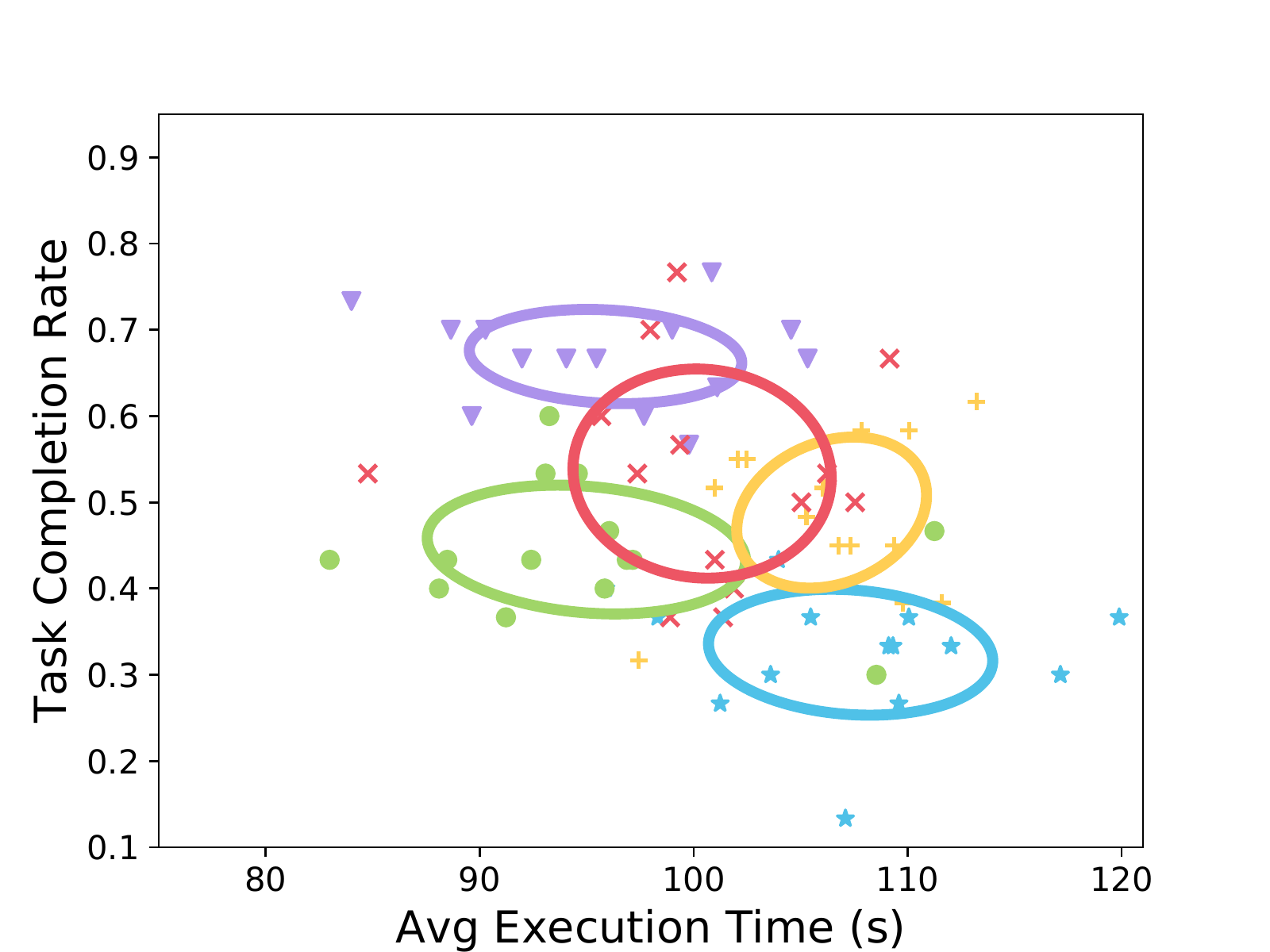}}\qquad \hspace{-2em}
    \subfloat[Hard]{\includegraphics[width=.32\linewidth]{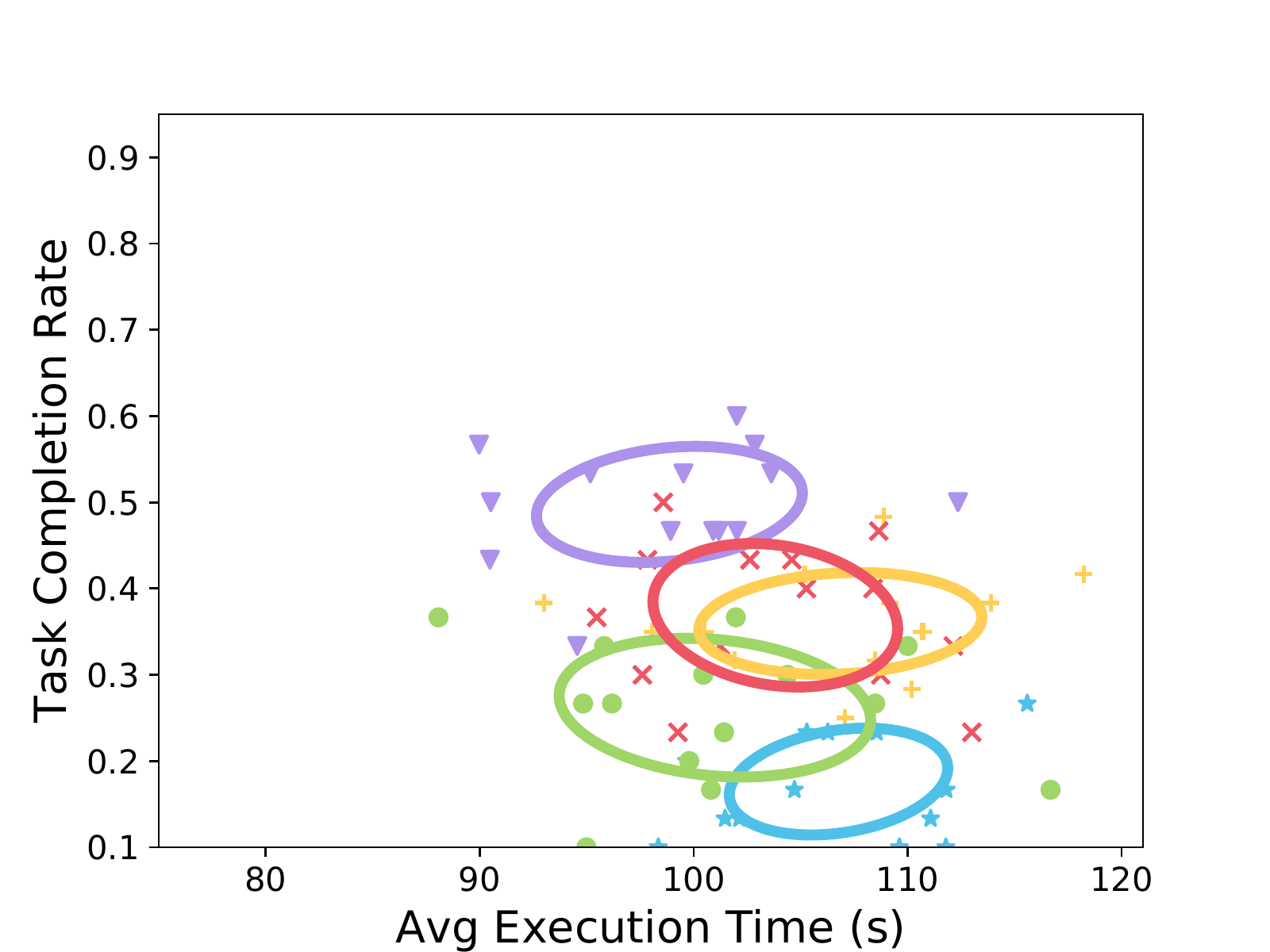}}
    \caption{Overall performances of GROP and four baseline methods in efficiency ($x$-axis) and task completion rate ($y$-axis). 
    Tasks are grouped based on their difficulties. 
    The ellipses represent the means and 2D standard variances of each approach. 
    GROP produced the highest task completion rate, while maintaining smaller or comparable execution time. 
    This observation is consistent over tasks of different difficulties. }
    \vspace{-1.7em}
    \label{fig:combined}
\end{figure*}

\vspace{.5em}
\noindent
\textbf{Baselines: }
GROP is evaluated through comparisons with the following baselines. 
All baselines are TAMP algorithms, and they vary in whether efficiency is considered in plan optimization, and whether feasibility is considered. 
All baselines select navigation goals by randomly sampling an obstacle-free position that is close to the unloading position. 

\begin{itemize}%[leftmargin=.5em]
    \item \textbf{Satisficing} (weakest baseline): Action costs are not considered, so it does not avoid long-distance navigation. 
    All actions share the same feasibility (FCN not used). 
    
    \item \textbf{PETLON}~\cite{lo2020petlon}: It considers plan efficiency, but does not quantitatively evaluate action feasibility. 
    
    \item \textbf{DVH}~\cite{driess2020deeph}: 
    It does not consider plan efficiency, but quantitatively evaluates action feasibility. 
    
    \item \textbf{FCN-Planning} (most competitive): The same as GROP except that the heatmap (Line~\ref{l:smp} in Algorithm~\ref{alg:GROP}) is not used for selecting 2D navigation goals. 
    
\end{itemize}

% \textbf{Satisficing} assumes that task level and motion level are not communicating with each other.
% \textbf{PETLON} corresponds to~\cite{lo2020petlon} where they predefine the state-mapping function and assume deterministic action outcomes during plan execution.
% \textbf{DVH} is similar to~\cite{driess2020deeph} because both methods use deep neural networks to learn the feasibility of motion-level behaviors (but~\cite{driess2020deeph} did not consider action efficiency).
It should be noted that, we cannot authentically implement DVH~\cite{driess2020deeph} for evaluation, because they used convolutional neural networks (CNNs) for task-level action feasibility evaluation, and we do not have a dataset from our domain for training the CNNs. 
% It should be noted that DVH only evaluates task-level action feasibility using a convolutional neural network (CNN), whereas GROP (ours) evaluates action feasibility at both task and motion levels. 
We did the best we could by replacing their CNN-based visual component with our FCN-based feasibility evaluator. 
% \textbf{FCN-Planning} is able to evaluate the feasibility value of symbolic actions but fails to use it to facilitate navigation behaviors (i.e., selecting 2D navigation goals).
% It is one ablative version of our method GROP.
%\xiaohan{Isn’t there some sort of sensible heuristic that could be compared? }

\vspace{.5em}
\noindent
\textbf{Experiment Setup: }
The mobile manipulator includes a UR5e robot arm, a Robotiq 2F-140 gripper, an RMP 110 mobile base, and a Velodyne VLP-16 lidar sensor. 
We used the Building-Wide Intelligence (BWI) codebase~\cite{khandelwal2017bwibots} to construct our simulation platform, which relies on the Gazebo physics engine~\cite{koenig2004design}.
We use a Rapidly exploring Random Tree (RRT) approach~\cite{lavalle1998rapidly} to compute motion-level manipulation plans. 
The navigation stack was built using the \texttt{move\_base} package of Robot Operating System (ROS)~\cite{quigley2009ros}.
% An example occupancy-grid map and a real-time trajectory computed by the path planner are shown in Fig.~\ref{fig:perception}. 
The robot's task planner is  ASP-based~\cite{gelfond2014knowledge,lifschitz2002answer} and we used the Clingo solver for computing task plans~\cite{gebser2014clingo}.

The dataset described in Section~\ref{sec:feasibility} was fed into an FCN for training $\Psi$.
We adapted the FCN-VGG16 model~\cite{long2015fully} and trained it with batch size $4$ and learning rate $e^{-3}$.
We used a machine equipped with an Intel 3.80GHz i7-10700k CPU and a GeForce RTX 3070 GPU on a Ubuntu system.

The test environment contains two tables, one for loading and the other (a long banquet table) for unloading. 
Obstacles (chairs) are randomly placed near the unloading table.
Positions and the number of chairs are dynamically changed for different environments.
An RGB camera is attached to the ceiling to capture  overhead images of environments. 
% We generated 30 environments in total for evaluation purposes. 
%We spawn different numbers of chairs (15, 20, and 25) to refer to difficulties of environment as being easy, normal, and hard. 
%For each difficulty, we generated 10 environments.
%We conducted 20 different unloading tasks for the robot in each environment.
% The task has been discussed in section~\ref{sec:problem_statement} where 
A mobile manipulator is tasked with moving three objects from the loading table to three different positions on the unloading table, where the robot can hold multiple objects at the same time. 
% to load $N$ objects from the loading table and unload them to $N$ different target positions (randomly sampled) on top of the long banquet table.
% We set the number of the objects $N$ to 3 and the unloading acceptable range $\alpha$ to 0.1 meters in the experiments.
There is a tolerance of $0.1m$ for unloading actions, and an unloading action is considered unsuccessful if the object is more than this distance away from the specified unloading position. 
% Unloading an object (out of $N$ objects) in the range of $\alpha$ is considered to be a subgoal completion.
Task completion is evaluated based on whether each ``seat'' of the table is set up. 
% Target object not in the range or robot getting in touch with obstacles are considered to be subgoal failures.
Reward $\mathcal{R}$ has a value of 40 in our utility function defined in Equation~\ref{eqn:utility}.

%\shiqi{Descirption of the three trajectories should be moved to the figure caption. }

\vspace{.5em}
\noindent 
\textbf{GROP vs. Baselines: }
Fig.~\ref{fig:combined} shows the main results from experiments of comparing GROP to the four baselines. 
There were a total of 420 different tasks in 30 different environments. 
% Five strategies (four baselines and GROP) have been quantitatively evaluated.
Each data point in the figure represents an average of 10 tasks. 
%\shiqi{Figure 5 does not have any point. There must be something wrong. }
% $x$-axis is the average execution time of completing one task, and $y$-axis is the task completion rate. 
We grouped the tasks based on their difficulties: Easy, Normal, and Hard. 
A task's \emph{difficulty} is measured by the total area that a robot can navigate to and unload an object from. 
For instance, a task with all unloading positions being surrounded by obstacles has a high difficulty. 
After sorting the tasks based on their difficulties, we evenly placed them into the three groups. 

GROP consistently performed better in task completion rate ($y$-axis) in all three settings, while maintaining high plan efficiency ($x$-axis). 
We also see that GROP performed particularly well in hard tasks where it produced the highest completion rate and the lowest action costs.  
%Our proposed method GROP produced a much higher task-completion accuracy while maintained a low total action cost. 
While PETLON generated efficient plans (comparable to GROP), it does not reason about feasibility, resulting in low completion rate. 
%\shiqi{Not sure which figure you are talking about here.. Will come back to it after things are more clear. }
DVH generates feasible plans (like FCN-Planning), but it does not consider action costs, resulting in long execution time in task completions. 
Results support our hypothesis that GROP improves plan efficiency without introducing additional action costs. 

\begin{table}[t]
%\tiny
\scriptsize
\centering
\vspace{.5em}
\caption{Task completion rate / average execution time in one of the environments with different robot's navigation velocities.}
\vspace{-.2em}
\resizebox{0.48\textwidth}{!}{%
\begin{tabular}{@{}lccc@{}}
\toprule
       & \emph{GROP}          & \emph{PETLON}        & \emph{DVH}           \\ \midrule
\emph{Slow}   & 0.80 / 166.16 & 0.63 / 166.46 & 0.73 / 204.04 \\ 
\emph{Medium} & 0.82 / 95.42  & 0.63 / 93.23  & 0.73 / 112.02 \\ 
\emph{Fast}   & 0.88 / 59.56  & 0.63 / 56.62  & 0.73 / 66.01  \\ \bottomrule
\end{tabular}%
}
\vspace{-2em}
\label{tb:speed}
\end{table}

\begin{figure*}[t]
\begin{center}
    % \vspace{-.5em}
    % \includegraphics[width = 17.5em,height=9em]{Figures/objects.png}
    \includegraphics[width=.95\textwidth]{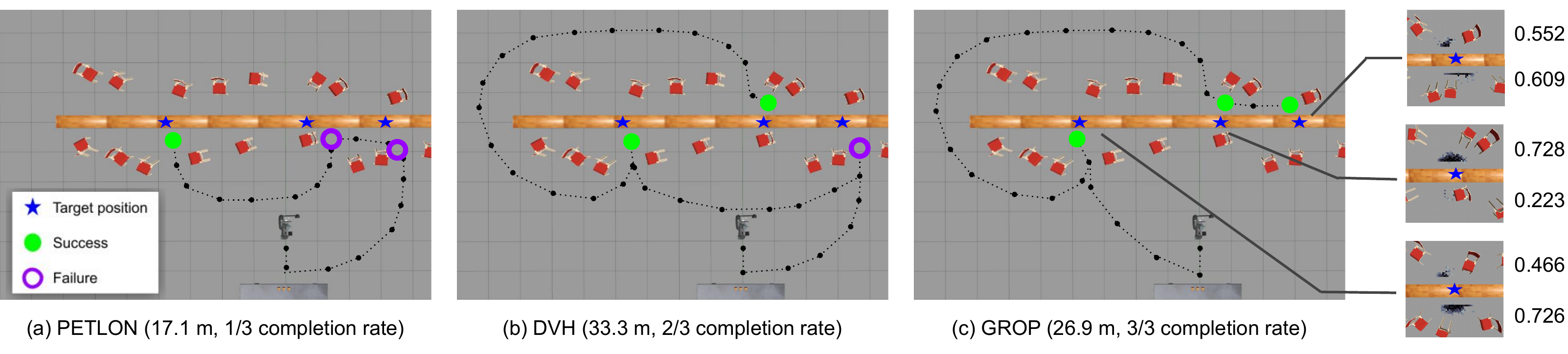}
    \vspace{-.5em}
    \caption{Three illustrative trials using GROP and two baselines (PETLON and DVH). 
    The robot needs to move three objects from the loading table (bottom) to three unloading target positions marked by {\color{blue}blue stars}, where the robot can hold multiple objects.
    {\color{green} Green dots} (or {\color{purple} purple circles}) represent a robot successfully (or unsuccessfully) navigating to the position and unloading an object to the corresponding target position. 
    Three heatmaps are overlaid onto overhead images, as shown on the right, indicating the feasibility values of navigating to and unloading from different positions. 
    The numbers on the very right represent task-level action feasibility values of unloading from one side of the table. 
    Under each subfigure, we present the total navigation distance and task completion rate, where we see GROP produced the highest completion rate, and performed better than DVH in efficiency.}
    % PETLON generated a plan of length 17.1 meters and succeeded 1 unloading task during execution.
    % DVH generated a plan of length 33.3 meters and succeeded 2 unloading task during execution.
    % GROP (ours) generated a plan of length 26.9 meters and succeeded the whole task during execution.
    % In addition, we show three (combined) heatmaps and their evaluated task-level feasibility using GROP.}
    % \caption{Everyday objects used in this research~\cite{thomason2016learning}, and the robot arm used for performing exploratory actions (grasping an object). }
    \vspace{-2em}
    \label{fig:grop_heatmap}
\end{center}
\end{figure*}

\vspace{.5em}
\noindent 
\textbf{Robot Velocities: }
In this experiment, we used three robots that move at different velocities: 0.2 m/s (\emph{Slow}), 0.4 m/s (\emph{Normal}), and 0.8 m/s (\emph{Fast}). 
Results are shown in Table~\ref{tb:speed}. 
Here we compare GROP to only the two baselines that are available from the literature (PETLON and DVH). 
We see that GROP outperforms the two baselines in task completion rate.
What is interesting is that when the robot moves fast, GROP automatically weighted feasibility more, because the robot will not take too long to complete a navigation task anyway. 
As a result, GROP produced the highest task completion rate of $0.88$ on a fast robot, while the baselines are not adaptive to the robot's velocity. 
% the action cost will be considered lower by GROP, thus GROP will focus more on task completions that results in a 88\% completion rate.
% When the robot is slower, GROP will consider more on plan efficiency that results in a lower execution time (lower than PETLON) but sacrifice some accuracies (80\%).

\vspace{.5em}
\noindent 
\textbf{Illustrative Trials in Simulation: }
Fig.~\ref{fig:grop_heatmap} shows three illustrative trials using GROP (ours) and two baselines (PETLON and DVH), where GROP produced the highest completion rate (3/3), while the baselines succeeded in at most two tasks. 
PETLON does not evaluate plan feasibility, and planned to unload objects to the middle and right positions from the south. 
In particular, unloading to the middle unloading position from the south is very difficult (with a feasibility value of $0.223$). 
PETLON does not take such factors into consideration, which produced failures in unloading to the middle position. 
DVH does not consider efficiency in plan optimization, and generated a plan with long-distance navigation actions. 
GROP incorporates both efficiency and feasibility, and produced the best overall performance. 

% Human gave the robot a task to unload three wooden boxes to three target positions (as shown in Fig.~\ref{fig:paths}) respectively.
% GROP agent took six partial views (there was a north and south side for each target position) and predicted heatmaps accordingly (Fig.~\ref{fig:grop_heatmap}). 
% Evaluated spatial feasibilities for partial views are listed in Table~\ref{tab:feasibility}.
% After considering both feasibility and efficiency, GROP generated a plan which successfully completed the whole, as shown in Fig.~\ref{fig:paths}(c).

% In comparison, PETLON agent (Fig.~\ref{fig:paths}(a)) selected the most efficient plan but failed at the middle and right unloading tasks because it did not evaluate action feasibility.
% Specifically, when unloading the object to the middle target position, it was hardly feasible but PETLON still insisted to unload from the south side.
% DVH agent selected the most feasible plan but failed at the right unloading task because it did not use the evaluated feasibility to select navigation goals.
% The right unloading position was selected too far from the table for the robot to perform unloading behavior. 

\vspace{.5em}
\noindent 
\textbf{Real Robot Demonstration: }
We demonstrate two trials of $T_1$ and $T_2$ using GROP on a real-robot platform. 
Instead of using a mobile manipulator, we used a robot system that includes two robots of a Segway-based mobile platform and a UR5e robot arm. 
% The task is similar to the one in simulation:
The mobile robot started from an initial position, and was tasked with loading a distant object (an orange cube in our case) from the arm robot.
The object was on the same table as the arm robot is, where the arm robot could pick the object, and place it onto the mobile robot to complete a loading behavior. 
%We manually constructed a simulation environment that is as similar as possible to the real-world configuration, and captured top-down view images for evaluating action feasibility. 
% We constructed similar scenarios in the simulation platform and used $\Psi$ to evaluate feasibility and sample navigation goals. 

\begin{figure}[t]
\vspace{.7em}
\begin{center}
    \includegraphics[width=0.8\columnwidth]{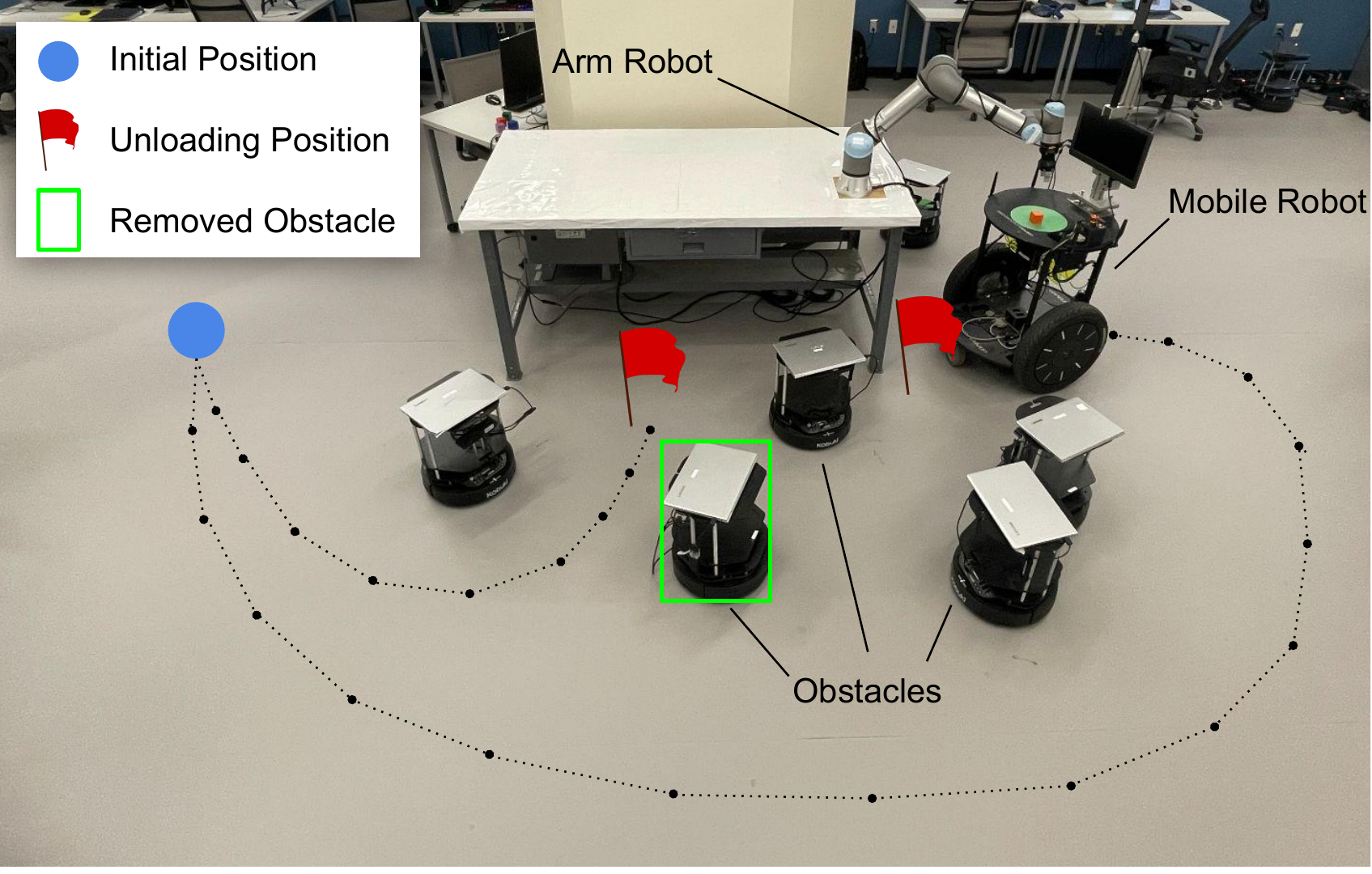}
    \vspace{-.5em}
    \caption{The arm robot is placing an object onto the mobile robot in trial $T_1$. 
    There are two loading positions on the south and east sides of the table, marked by {\color{red}red flags}.
    The mobile robot's initial position is shown as the {\color{blue}blue dot}. 
    The {\color{green}green box} highlights the obstacle that was removed in $T_2$. } 
    \label{fig:real}
    \vspace{-2em}
\end{center}
\end{figure}

In trial $T_1$, the system computed the task-level feasibility values of loading from the south and east sides: $0.377$ and $0.721$. 
The corresponding costs were $7.5$ and $19.3$ respectively (distances of $4.5 m$ and $11.6 m$), where the robot moved at speed $0.6 m/s$. 
GROP evaluated the utility values ($7.6$ and $9.5$ in this case), and decided to load from the east side (less efficient but more feasible), as shown in Fig.~\ref{fig:real}.

In trial $T_2$, the robot system worked on the same task, while one obstacle (green box in Fig.~\ref{fig:real}) was removed from the environment. 
The obstacle removal changed the feasibility: Loading from the south has higher feasibility of $0.520$, and a higher utility of $13.3$. 
Accordingly, the mobile robot decided to load the object from the south, where there existed little chance of failing in the loading behavior but the overall efficiency was significantly improved. 
% After trading-off plan feasibility and efficiency, the system decided to load from the south which is a bit risky but cost-efficient.
In both demonstration trials, the robot system succeeded in loading the object to the mobile platform. 

%\shiqi{Looks like this subsection is incomplete. }

%We set up different difficulties (by controlling the number and position of chairs) of the environment
%Two different plans are selected in the demonstrated two trials due 

\section{Conclusion and Future Work}
%In this work, we focus on perception-based task and motion planning problems for mobile manipulators.
This paper introduces an algorithm, called \textbf{G}rounded \textbf{RO}bot Task and Motion \textbf{P}lanning (GROP), that considers both efficiency and feasibility for robot task-motion planning. 
GROP visually grounds spatial relationships to probabilistically evaluate action feasibility, and is particularly suitable for TAMP domains with long-term robot operations (e.g., long-distance navigation). 
We have extensively evaluated GROP in simulation using a mobile manipulator, and demonstrated it using a real robot system that includes a mobile robot and an arm robot. 
Results showed that GROP outperformed competitive baselines from the literature in plan efficiency without introducing additional action costs. 

In this paper, we empirically evaluated the performance of GROP, while there is room to improve the evaluation through formal analysis, e.g., about its completeness and optimality.
The difficulty comes from the different problem representations of task planning and motion planning.
Also, the FCN-based feasibility evaluator is data-driven, where formal analysis is difficult.
%This paper does not formally analyze properties of GROP, such as its completeness, soundness, and optimality. 
% Conducting such formal analysis is an important direction for future work. 
%\yifeng{This sentence sounds quite negative and might let reviewers think that we haven't completed our jobs well.}
The current implementation of GROP relies on top-down views. 
It would be interesting to investigate the feasibility of applying egocentric vision to GROP. 
Due to the various viewpoints, we expect GROP to require a greater amount training data in this setting. 
%\yifeng{"various viewpoints" sounds like a multi-view camera setup. I suspect we mean this by "ever-changing" viewpoint?}

\section*{Acknowledgements}
%This work has taken place at the Autonomous Intelligent Robotics (AIR) Group, SUNY Binghamton. 
%AIR research is supported in part by grants from the NSF (NRI-1925044), Ford, OPPO, and SUNY RF.
This work has taken place in the Autonomous Intelligent Robotics (AIR) group at SUNY Binghamton, and in the Learning Agents Research Group (LARG) at UT Austin. 
AIR research is supported in part by grants from NSF (IIS-1925044), Ford Motor Company, OPPO, and SUNY Research Foundation. 
LARG research is supported in part by NSF (CPS-1739964, IIS-1724157, FAIN-2019844), ONR (N00014-18-2243), ARO (W911NF-19-2-0333), DARPA, Lockheed Martin, GM, Bosch, and UT Austin's Good Systems grand challenge.  
Peter Stone serves as the Executive Director of Sony AI America and receives financial compensation for this work.  
The terms of this arrangement have been reviewed and approved by the University of Texas at Austin in accordance with its policy on objectivity in research.

\clearpage
\bibliographystyle{IEEEtran}
\bibliography{references.bib}

\end{document}